\documentclass[11pt,a4paper]{article}
\usepackage[hyperref]{style_files/acl2020}
\usepackage{times}
\usepackage{latexsym}

\usepackage{microtype}

\usepackage{url}

\usepackage{xspace}
\usepackage{xargs} %
\usepackage[table]{colortbl}

\usepackage{booktabs}

\usepackage{array}
\usepackage{microtype}
\usepackage{graphicx}
\usepackage[labelformat=simple]{subcaption}

\usepackage{amsmath}
\definecolor{OliveGreen}{cmyk}{0.64,0,0.95,0.40}
\usepackage{comment}

\usepackage{bbding}
\usepackage{pifont}
\usepackage{wasysym}
\usepackage{amssymb}
\usepackage{multirow}

\aclfinalcopy %
\def\aclpaperid{2330} %

\usepackage[colorinlistoftodos,prependcaption,textsize=tiny]{todonotes}

\usepackage{marginnote}

\newcommandx{\siva}[2][1=]{\todo[linecolor=red,backgroundcolor=red!10,bordercolor=red,#1]{SR: #2}\xspace}
\newcommand{\ignore}[1]{}

\newcommand{\secref}[2][]{\S#1\ref{#2}\xspace}
\newcommand{\figref}[2][]{Figure#1~\ref{#2}\xspace}
\newcommand{\tabref}[2][]{Table#1~\ref{#2}\xspace}

\newcommand{\dataset}[1]{\textit{#1}\xspace}

\newcommand{\gref}{\dataset{RefCOCOg}}
\newcommand{\grefe}{\dataset{Ref-Easy}}
\newcommand{\grefh}{\dataset{Ref-Hard}}
\newcommand{\grefa}{\dataset{Ref-Adv}}

\newcommand{\grefplus}{\dataset{RefCOCO+}}
\newcommand{\greforiginal}{\dataset{RefCOCO}}
\newcommand{\greftest}{\dataset{RefCOCOg-Test}}
\newcommand{\grefDev}{\dataset{RefCOCOg-Dev}}

\newcommand{\greft}{\dataset{Ref-Test}}
\newcommand{\grefd}{\dataset{Ref-Dev}}

\usepackage{bbding}
\usepackage{pifont}
\usepackage{wasysym}
\usepackage{amssymb}
\usepackage{balance}

\begin{document}

\title{Words aren't enough, their order matters:\\ On the Robustness of Grounding Visual Referring Expressions}
\author{ Arjun R. Akula\textsuperscript{{\rm 1}}\thanks{\  \ Work done in part while AA was intern at Amazon AI.}\ , Spandana Gella\textsuperscript{{\rm 2}},  Yaser Al-Onaizan\textsuperscript{{\rm 2}},  Song-Chun Zhu\textsuperscript{{\rm 1}},  Siva Reddy\textsuperscript{{\rm 3}} \\
\textsuperscript{\rm 1}UCLA Center for Vision, Cognition, Learning, and Autonomy, 
\textsuperscript{\rm 2}Amazon AI\\
\textsuperscript{\rm 3}Facebook CIFAR AI Chair, Mila; McGill University\\
\texttt{aakula@ucla.edu}, \texttt{sgella@amazon.com} \\
}

\maketitle

\begin{abstract}
Visual referring expression recognition is a challenging task that requires natural language understanding in the context of an image.
We critically examine \gref, a standard benchmark for this task, using a human study and show that 83.7\% of test instances do not require reasoning on linguistic structure, i.e., words are enough to identify the target object, the word order doesn't matter.
To measure the true progress of existing models, we split the test set into two sets, one which requires reasoning on linguistic structure and the other which doesn't.
Additionally, we create an out-of-distribution dataset \grefa by asking crowdworkers to perturb in-domain examples such that the target object changes.
Using these datasets, we empirically show that existing methods fail to exploit linguistic structure and are 12\% to 23\% lower in performance than the established progress for this task.
We also propose two methods, one based on contrastive learning and the other based on multi-task learning, to increase the robustness of ViLBERT, the current state-of-the-art model for this task. Our datasets are publicly available at \url{https://github.com/aws/aws-refcocog-adv}.
\end{abstract}

\section{Introduction}
\begin{figure}[t]
\centering
  \includegraphics[width=\linewidth]{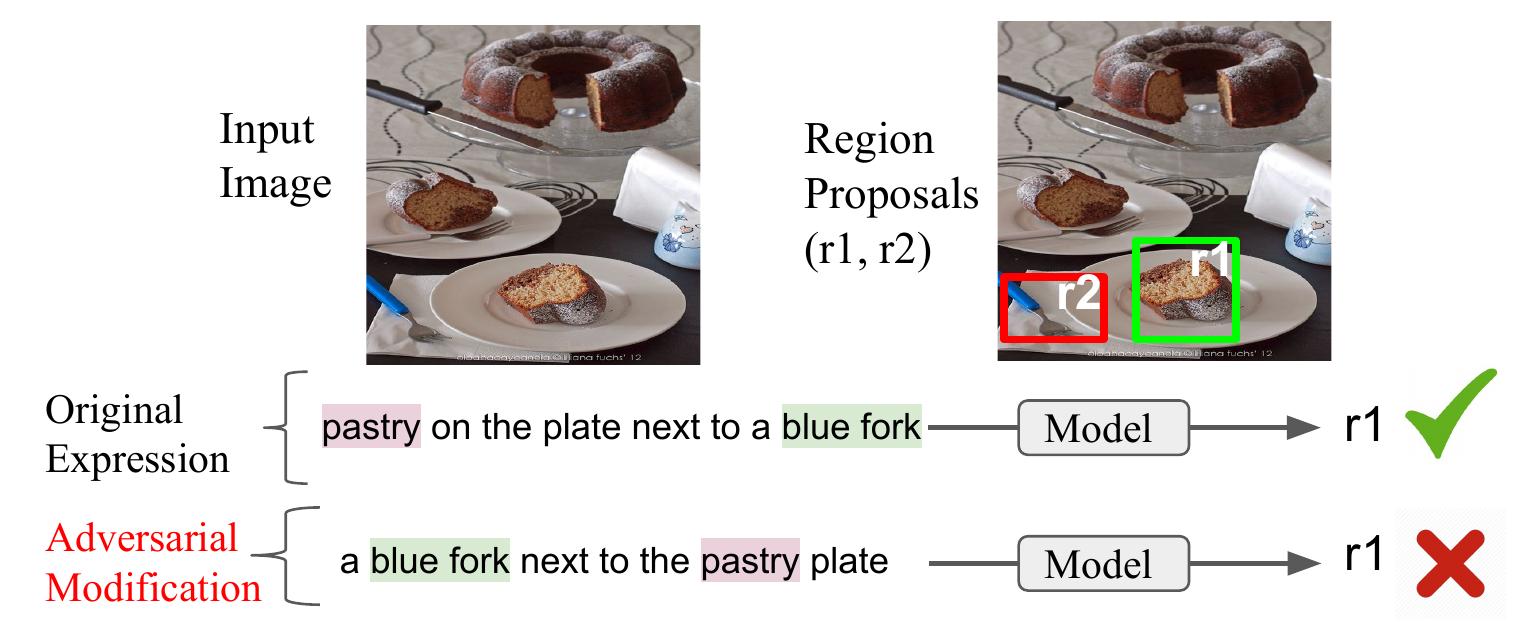}
  \vspace{-1.5em}
  \caption{An example of the visual referring expression recognition task. If the word \textit{pastry} is present in the referring expression, models prefer the bounding box \textit{r1} (highlighted in green) irrespective of the change in linguistic structure (word order).}~\label{fig:examples}
  \vspace{-2em}
\end{figure}

Visual referring expression recognition is the task of identifying the object in an image referred by a natural language expression \cite{kazemzadeh2014referitgame,NagarajaMD16,mao2016generation,hu2016natural}. 
\figref{fig:examples} shows an example. 
This task has drawn much attention due to its ability to test a model's understanding of natural language in the context of visual grounding and its application in downstream tasks such as image retrieval~\cite{young2014image} and question answering~\cite{antol2015vqa,zhu2016visual7w}. 
To track progress on this task, various datasets have been proposed, in which real world images are annotated by crowdsourced workers \cite{kazemzadeh2014referitgame,mao2016generation}.
Recently, neural models have achieved tremendous progress on these datasets \cite{yu2018mattnet,lu2019vilbert}.
However, multiple studies have suggested that these models could be exploiting strong biases in these datasets \cite{CirikMB18,liu2019clevrref}.
For example, models could be just selecting a salient object in an image or a referring expression without recourse to linguistic structure (see \figref{fig:examples}).
This defeats the true purpose of the task casting doubts on the actual progress.

In this work, we  examine \gref dataset \cite{mao2016generation}, a popular testbed for evaluating referring expression models, using crowdsourced workers.
We show that a large percentage of samples in the \gref test set indeed do not rely on linguistic structure (word order) of the expressions.
Accordingly, we split \gref test set into two splits, \grefe and \grefh, where linguistic structure is key for recognition in the latter but not the former (\secref{sec:ling_struct}).
In addition, we create a new out-of-distribution\footnote{This is a \textit{contrast set} according to \citet{gardner2020evaluating}} dataset called \grefa using \grefh by rewriting a referring expression such that the target object is different from the original annotation (\secref{sec:collection}).
We evaluate existing models on these splits %
and show that the true progress is at least 12-23\% behind the established progress, indicating there is ample room for improvement (\secref{sec:diganising-rer}).
We propose two new models, one which make use of contrastive learning using negative examples, and the other based on multi-task learning, and show that these are slightly more robust than the current state-of-the-art models (\secref{sec:models}).

\section{Importance of linguistic structure}
\label{sec:ling_struct}
\gref is the largest visual referring expression benchmark available for real world images \cite{mao2016generation}.
Unlike other referring expression datasets such as \textit{RefCOCO} and \textit{RefCOCO+}~\cite{kazemzadeh2014referitgame}, a special care has been taken such that expressions are longer and diverse. 
We therefore choose to examine the importance of linguistic structure in \gref.

\newcite{CirikMB18} observed that when the words in a referring expression are shuffled in random order, the performance of existing models on \gref drops only a little.
This suggests that models are relying heavily on the biases in the data than on linguistic structure, i.e., the actual sequence of words.
Ideally, we want to test models on samples where there is correlation between linguistic structure and spatial relations of objects, and any obscurity in the structure should lead to ambiguity.
To filter out such set, we use humans.

We randomly shuffle words in a referring expression to distort its linguistic structure, and ask humans to identify the target object of interest via predefined bounding boxes. %
Each image in \gref test set is annotated by five Amazon Mechanical Turk (AMT) workers and when at least three annotators select a bounding box that has high overlap with the ground truth, we treat it as a correct prediction.
Following \newcite{mao2016generation}, we set 0.5 IoU (intersection over union) as the threshold for high overlap.
Given that there are at least two objects in each image, the optimal performance of a random choice is less than 50\%.\footnote{On average, there are 8.2 bounding boxes per image.}
However, we observe that human accuracy on distorted examples is 83.7\%, indicating that a large portion of \gref test set is insensitive to linguistic structure. 
Based on this observation, we divide the test set into two splits for fine-grained evaluation of models: \textbf{\grefe} contains samples insensitive to linguistic structure and \textbf{\grefh} contains sensitive samples (statistics of the splits are shown in \tabref{tab:stats}).

\begin{table}[t]
\footnotesize
    \centering
    \tabcolsep 1.5pt
    \begin{tabular}{lccc}
    \toprule
    & \grefe & \grefh & \grefa \\
    \midrule
data size & $\underset{(83.7\% \textrm{ of \gref})}{8034}$ & $\underset{(16.3\% \textrm{ of \gref})}{1568}$ & $3704$ \\
& & & \\[-0.5em]
 $\underset{\textrm{in\; words}}{\textrm{avg. length}}$ & $8.0$ &  $10.2$ &  $11.4$ \\
\bottomrule
    \end{tabular}
    \caption{Statistics of \grefe, \grefh and \grefa. \grefe and \grefh indicate the proportion of samples in \gref test set that are insensitive and sensitive to linguistic structure respectively. 
    }
    \label{tab:stats}
    \vspace{-1em}
\end{table}

\section{An out-of-distribution dataset}\label{sec:collection}

Due to unintended annotation artifacts in \gref, it is still possible that models could perform well on \grefh without having to rely on linguistic structure, e.g., by selecting frequent objects seen during training time.
Essentially, \grefh is an in-distribution split.
To avoid this, we create \textbf{\grefa}, an adversarial test set with samples that may be fall out of training distribution.

\ignore{
\begin{table*}[t]
\tabcolsep 2pt
\begin{center}
\begin{tabular}{p{5cm}p{5cm}p{5cm}}
{\textcolor{blue}{\bf{\grefe Examples}}} & {\textcolor{blue}{\bf{\grefh Examples}}} & {\textcolor{blue}{\bf{\grefa Examples}}}\\
   \includegraphics[width=0.5\columnwidth,height=0.25\columnwidth]{images/23314E} &
 \includegraphics[width=0.5\columnwidth,height=0.25\columnwidth]{images/23314E} & \includegraphics[width=0.5\columnwidth,height=0.25\columnwidth]{images/23314A}\\
 \small{A green bush in a black pot} & \small{The green plant is beside the girl} & \small{A girl with a rolling suitcase is walking past the green plant} \\
   \includegraphics[width=0.5\columnwidth,height=0.25\columnwidth]{images/323040E} & \includegraphics[width=0.5\columnwidth,height=0.25\columnwidth]{images/323040E} & \includegraphics[width=0.5\columnwidth,height=0.25\columnwidth]{images/323040A}\\
\small{Black Iphone} & \small{Telephone underneath a tablet} & \small{A black and white e-reader next to a telephone} \\
   \includegraphics[width=0.5\columnwidth,height=0.25\columnwidth]{images/1969140E} &
 \includegraphics[width=0.5\columnwidth,height=0.25\columnwidth]{images/1969140E} & \includegraphics[width=0.5\columnwidth,height=0.25\columnwidth]{images/1969140A}\\
\small{3D wallpaper inside room} & \small{White color painting on the wall} & \small{Turned off color TV to the right of the painting on the wall} \\
 \end{tabular}
\caption{Random examples from  \grefe, \grefh, and \grefa splits.}
\label{tab:examples}
\end{center}
\end{table*}
}

\definecolor{darkgreen}{rgb}{0.0, 0.5, 0.13}

\begin{figure*}[t]
\footnotesize
\tabcolsep 1.2pt
\hspace{-1em}\begin{tabular}{c p{5.2cm}}
\multirow{5}{*}{\includegraphics[width=2.9cm,height=7em]{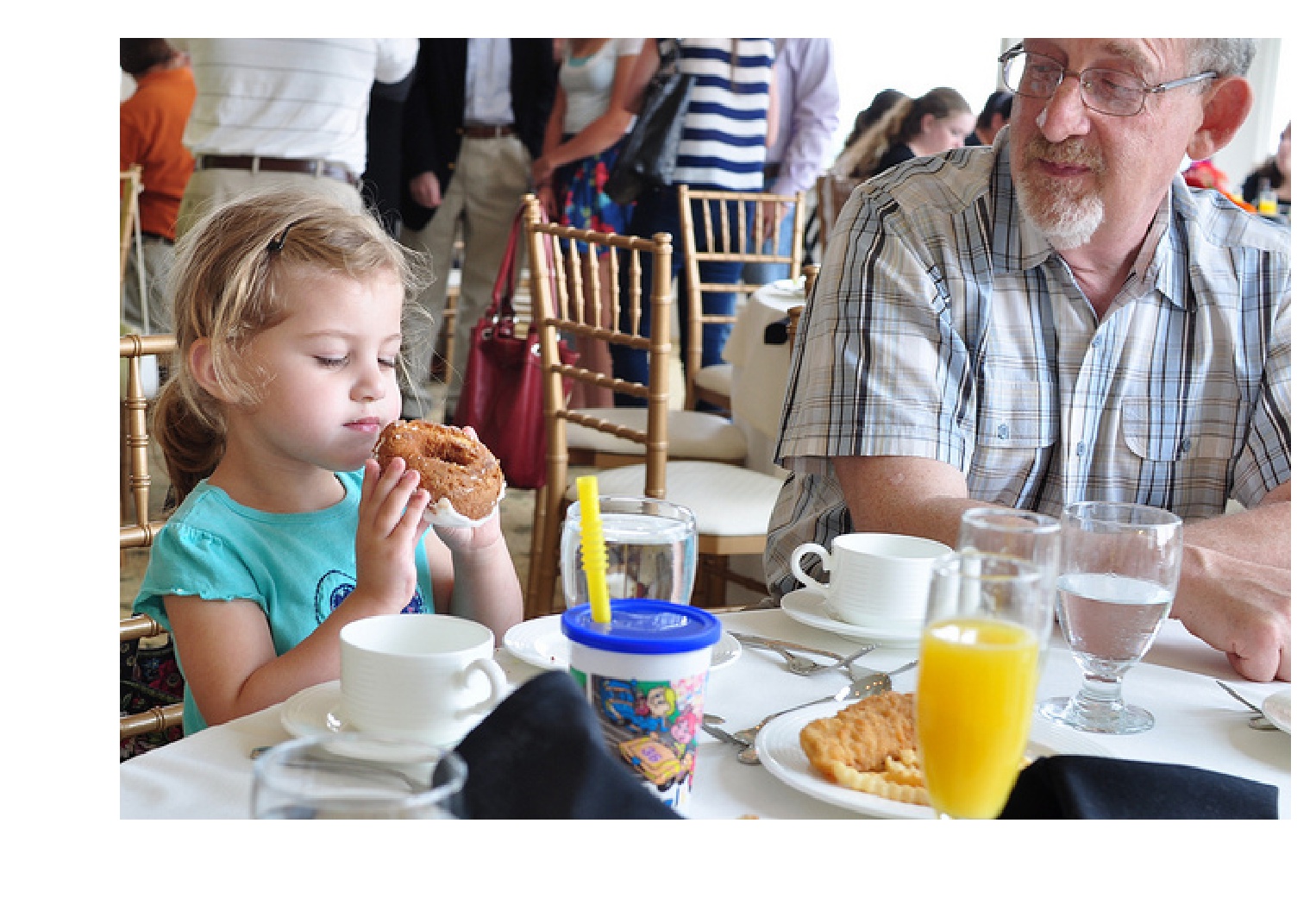}} & \textcolor{darkgreen}{\textbf{Easy:}}  A dinning table with cake and drinks \\
 & \textcolor{blue}{\textbf{Hard:}} A chair with a purse hanging from it  \\
  & \textcolor{red}{\textbf{Adv:}} The purse which is hanging from a chair \\  
\multirow{5}{*}{\includegraphics[trim=2em 0 0 13em,clip,height=6.5em,width=2.9cm]{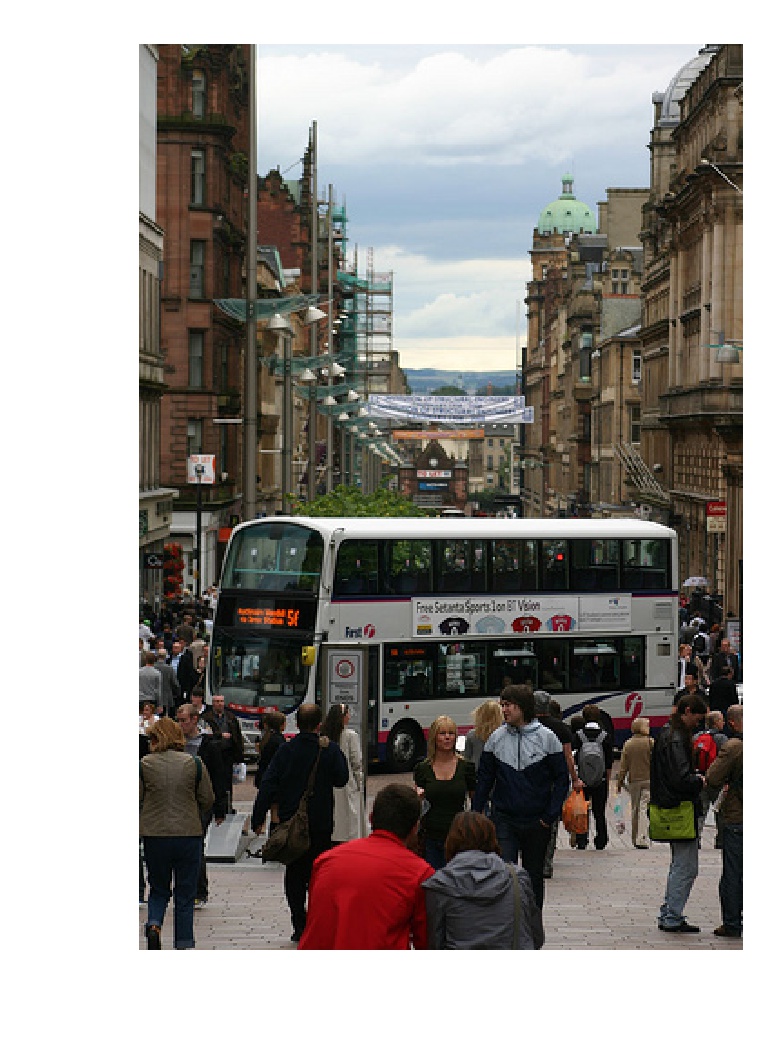}} & \textcolor{darkgreen}{\textbf{Easy:}}  Bus \\
 & \textcolor{blue}{\textbf{Hard:}} Bus in the middle of the crowd  \\
  & \textcolor{red}{\textbf{Adv:}} The crowd that the bus is in the middle of \\
  \\
\multirow{5}{*}{\includegraphics[height=7em]{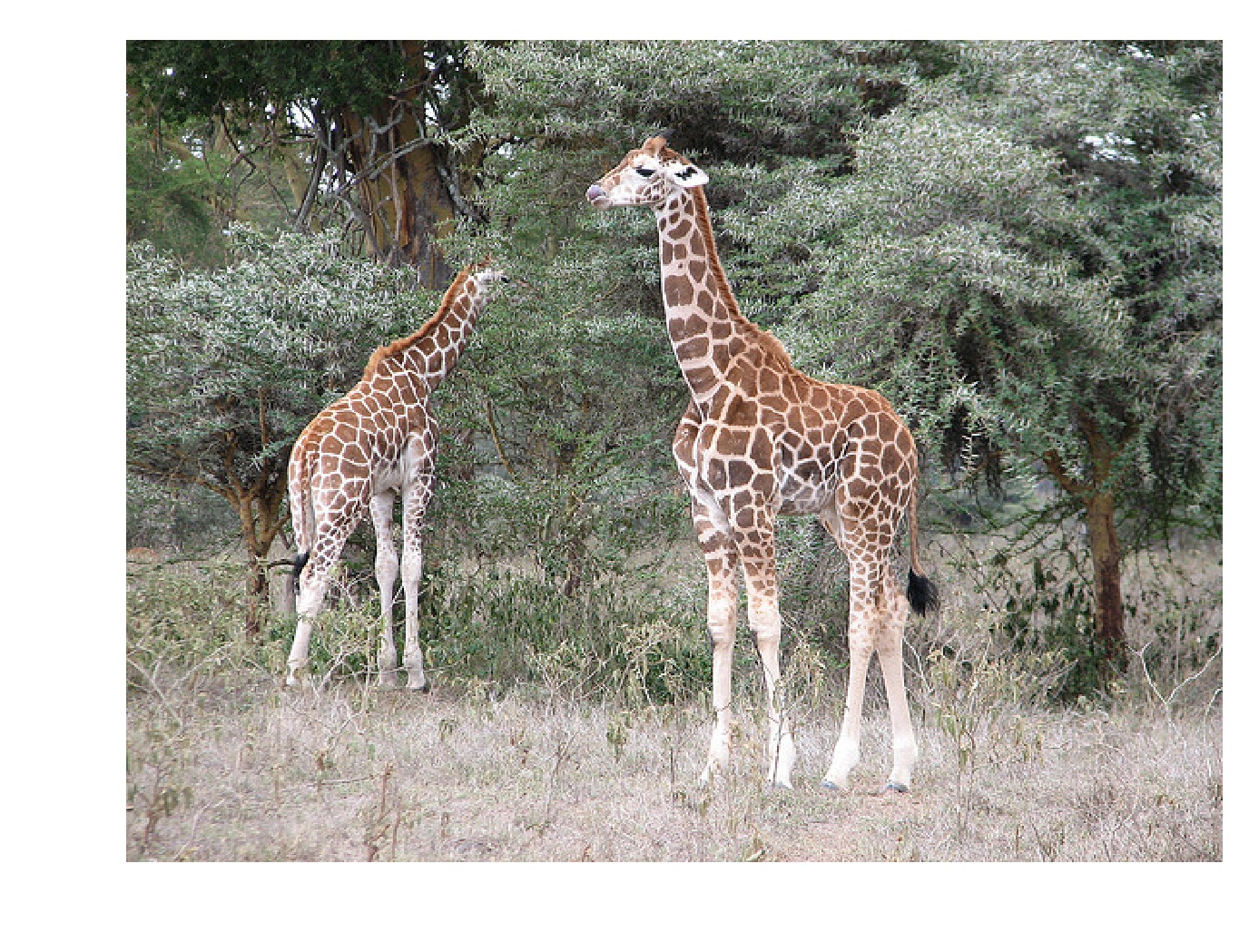}}  & \textcolor{darkgreen}{\textbf{Easy:}}   The larger of two giraffes \\
 & \textcolor{blue}{\textbf{Hard:}}  A giraffe eating leaves off the tree \\
  & \textcolor{red}{\textbf{Adv:}} The giraffe that is not eating leaves off the tree \\  
 \\
\multirow{5}{*}{\includegraphics[width=2.9cm,height=8em]{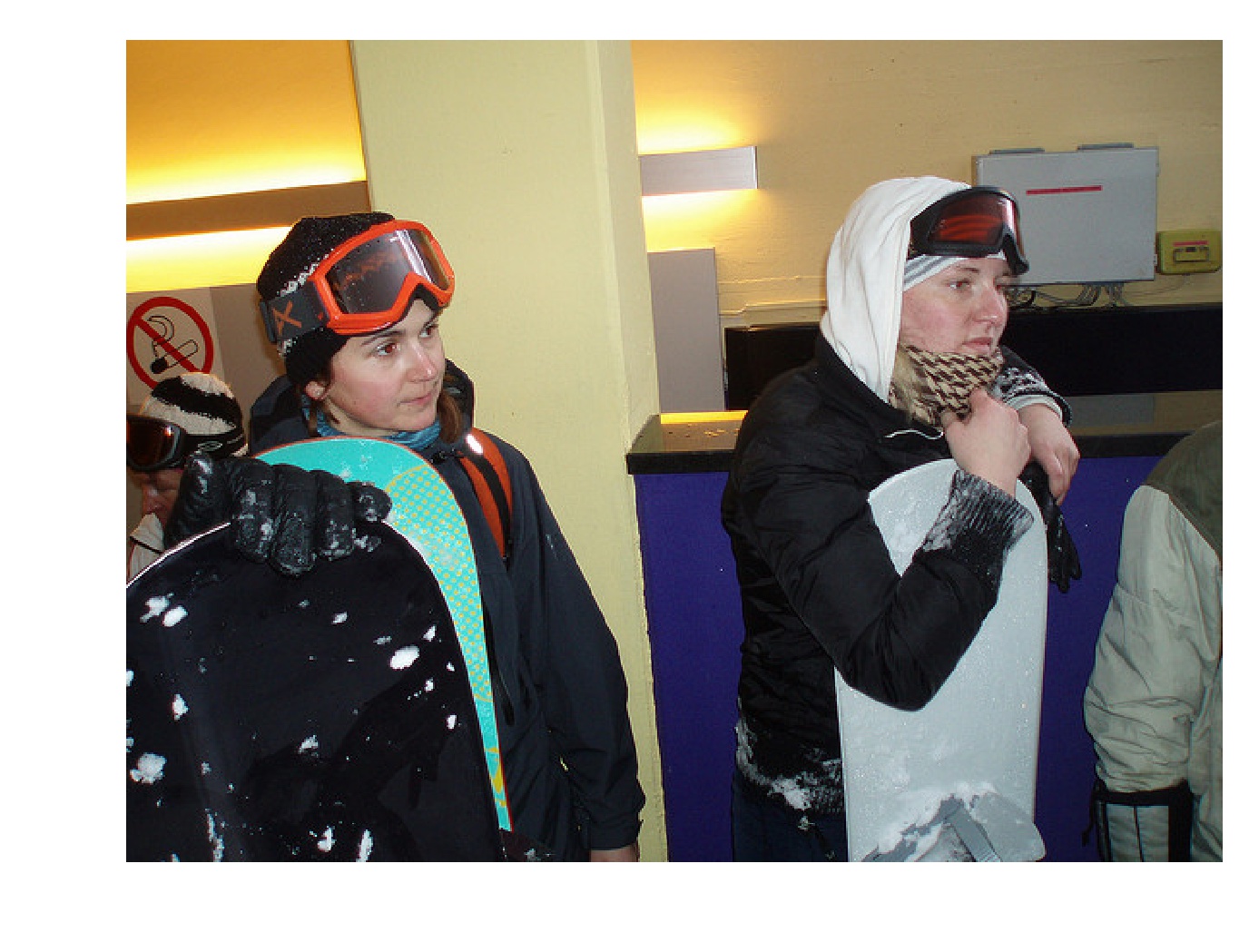}}  & \textcolor{darkgreen}{\textbf{Easy:}}  A blue snowboard \\
 & \textcolor{blue}{\textbf{Hard:}} A woman wearing a blue jacket and orange glasses next to a woman with a white hood \\
  & \textcolor{red}{\textbf{Adv:}} A woman with a white hood, next to a woman wearing orange glasses and a blue jacket. \\  
\\
\end{tabular}
\begin{tabular}{c p{5.2cm}}
\multirow{5}{*}{\includegraphics[height=7em]{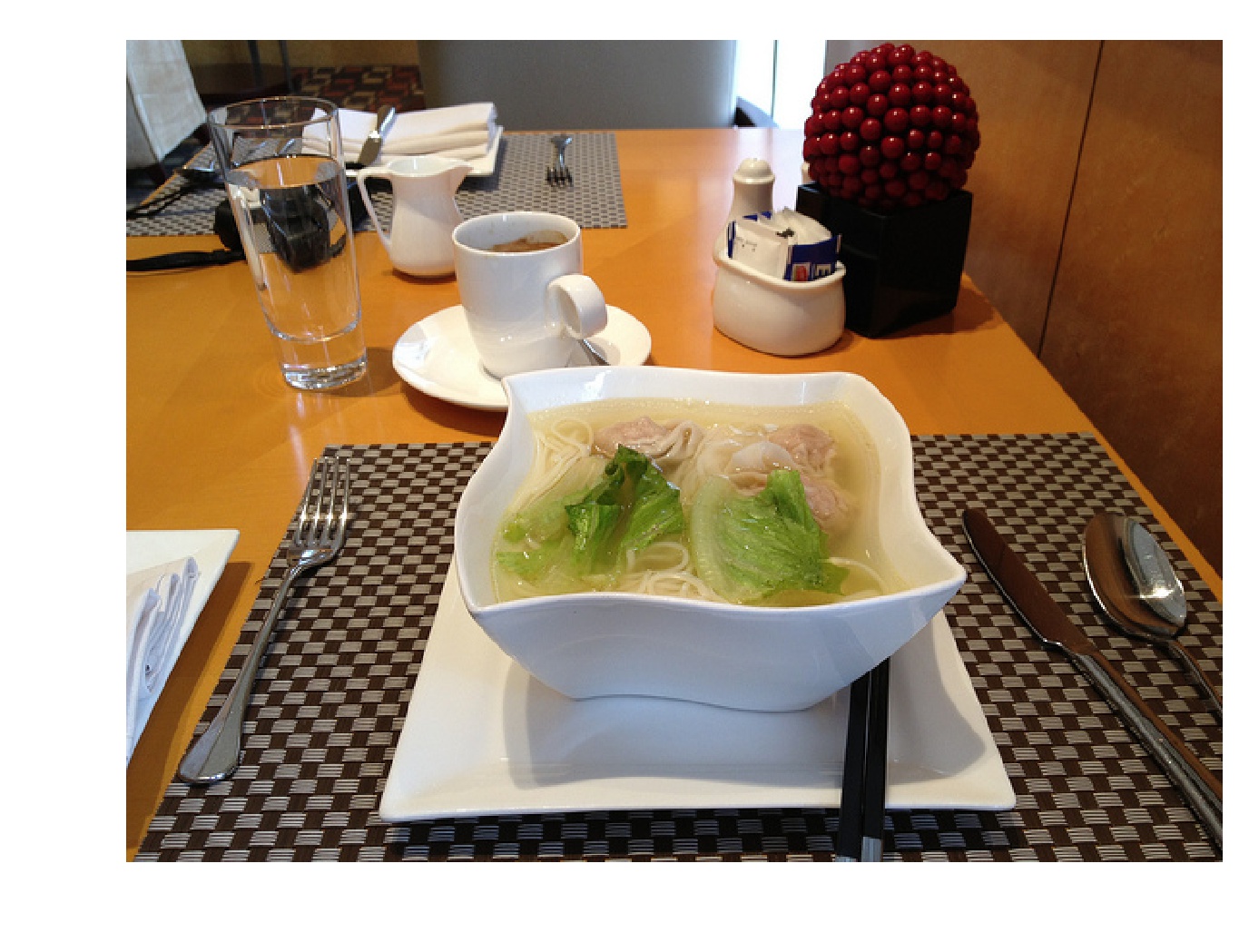}}  &  \textcolor{darkgreen}{\textbf{Easy:}}  Water in a tall, clear glass \\
 & \textcolor{blue}{\textbf{Hard:}} The glass of water next to the saucer with the cup on it \\
  & \textcolor{red}{\textbf{Adv:}} The cup on the saucer, next to the glass of water \\  
\multirow{5}{*}{\includegraphics[height=7em]{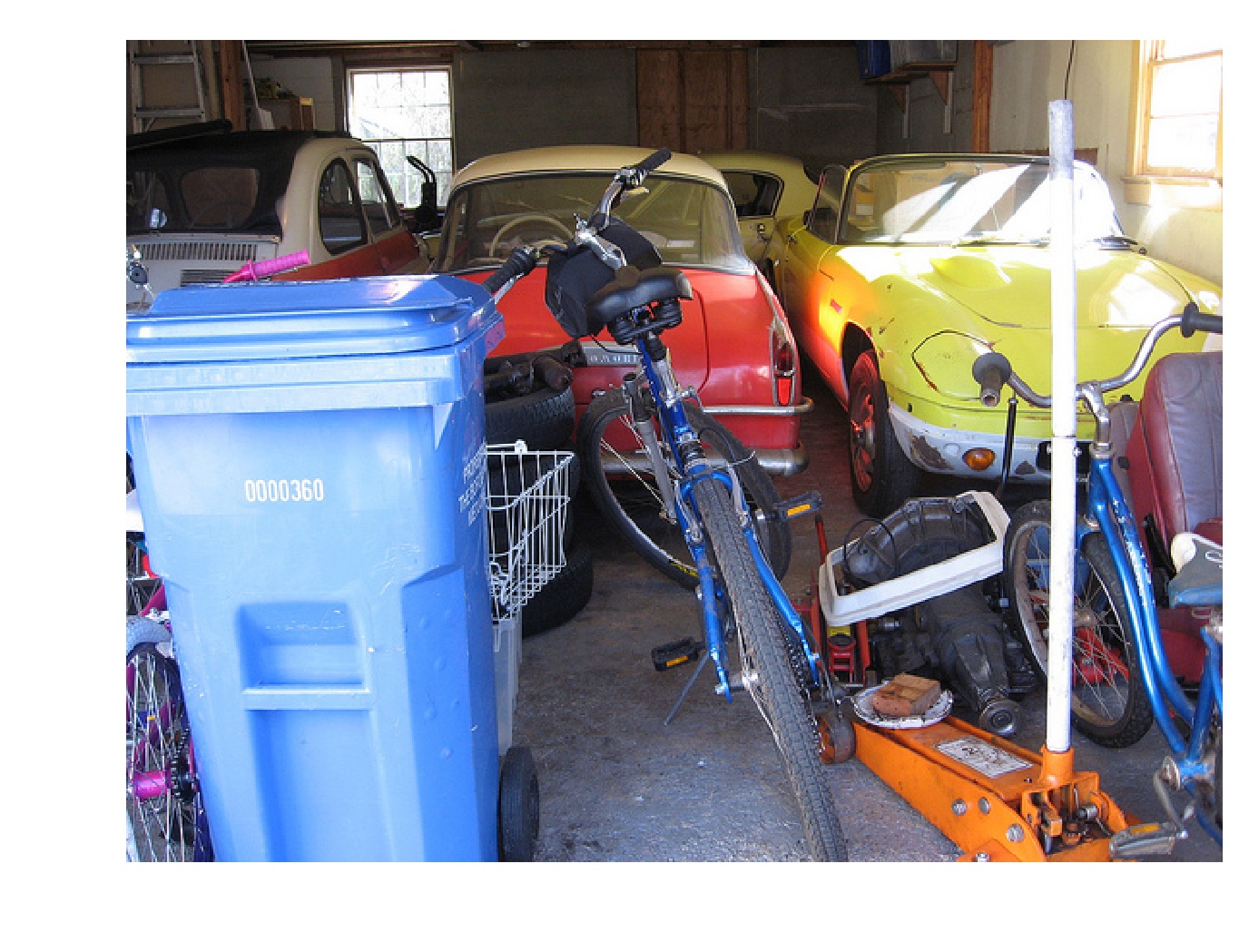}} \\
& \textcolor{darkgreen}{\textbf{Easy:}}     The short blue bike on the right \\
 & \textcolor{blue}{\textbf{Hard:}}  The blue bike behind the red car \\
  & \textcolor{red}{\textbf{Adv:}} The red car behind the blue bike \\  
 \\
 \\
\multirow{5}{*}{\includegraphics[trim=0 0 0 0,clip,height=6.5em,width=2.9cm]{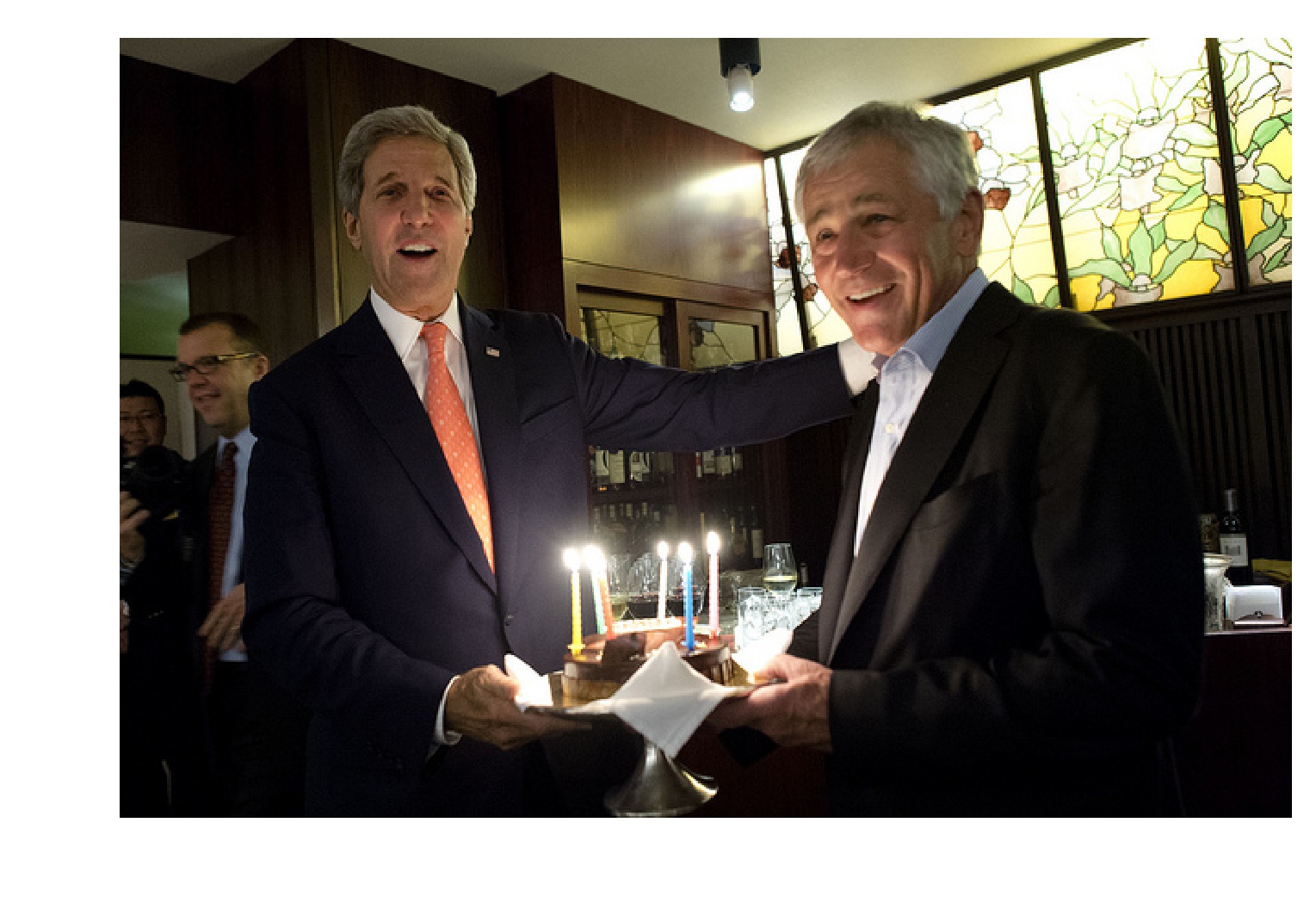}} & \textcolor{darkgreen}{\textbf{Easy:}}   The man with the glasses on \\
 & \textcolor{blue}{\textbf{Hard:}} A man holding a cake that is not wearing a tie \\
  & \textcolor{red}{\textbf{Adv:}} The man holding a cake that is wearing a tie \\  
\\
\multirow{5}{*}{\includegraphics[trim=0 0 0 5em,clip,width=2.9cm,height=7em]{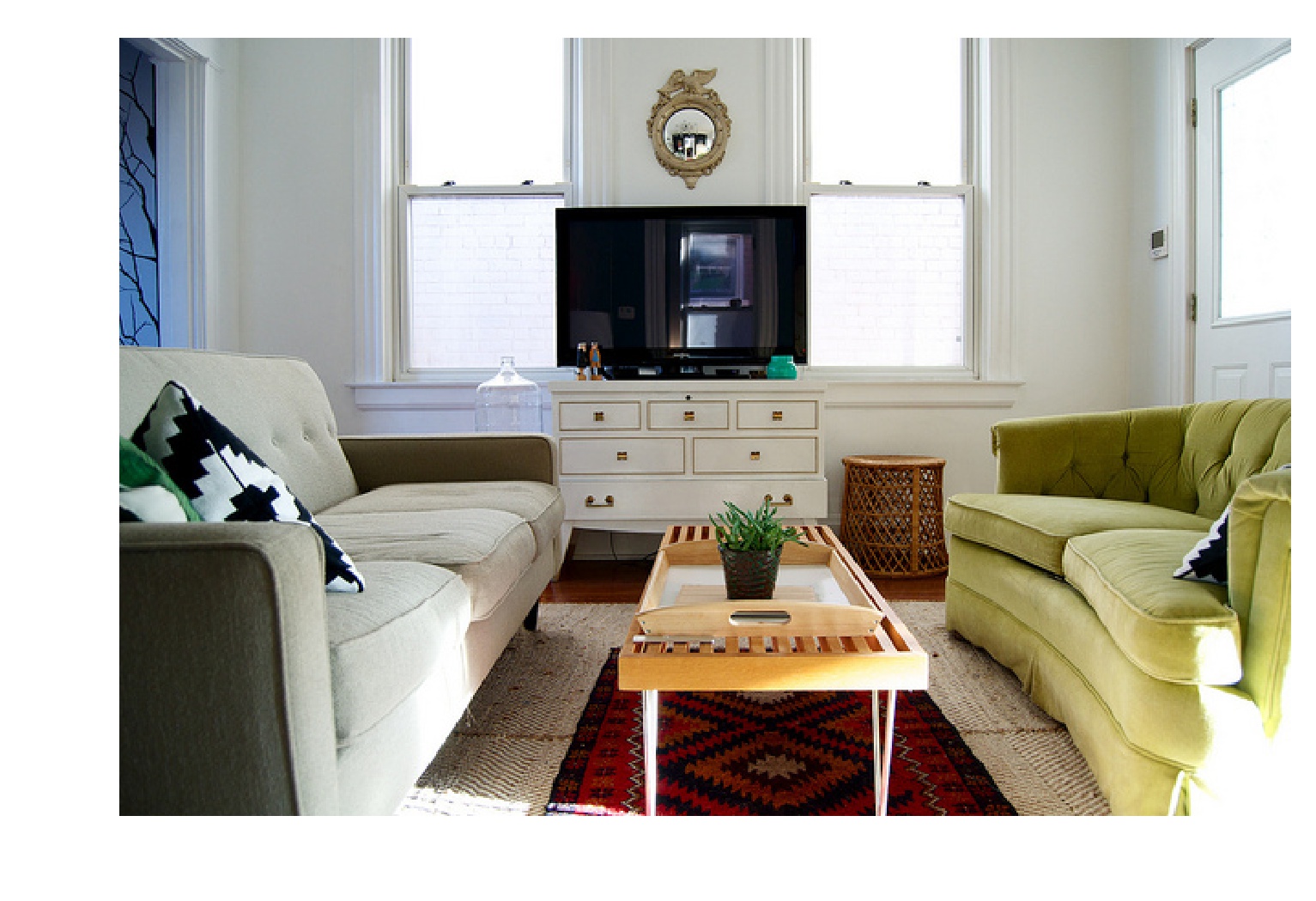}}  & \textcolor{darkgreen}{\textbf{Easy:}}   A green cushion couch with a pillow \\
 & \textcolor{blue}{\textbf{Hard:}} A green couch across from a white couch \\
  & \textcolor{red}{\textbf{Adv:}} A white couch across from a green couch \\  
  \\
\end{tabular}
\vspace{-1em}
\caption{Examples from  \grefe, \grefh, and \grefa splits. As seen, \grefh and \grefa have several words in common but differ in their linguistic structure and the target object of interest.}
\vspace{-1em}
\label{tab:examples}
\end{figure*}

We take each sample in \grefh and collect additional referring expressions such that the target object is different from the original object.
We chose the target objects which humans are most confused with when the referring expression is shuffled (as described in the previous section).
For each target object, we ask three AMT workers to write a referring expression while retaining most content words in the original referring expression.
In contrast to the original expression, the modified expression mainly differs in terms of the structure while sharing several words.
For example, in \figref{fig:examples}, the adversarial sample is created by swapping \textit{pastry} and \textit{blue fork} and making \textit{plate} as the head of \textit{pastry}.
We perform an extra validation step to filter out bad referring expressions.
In this step, three additional AMT workers select a bounding box to identify the target object, and we only select the samples where at least two workers achieve IoU $>$ 0.5 with the target object.

Since the samples in \grefa mainly differ in linguistic structure with respect to \grefh, we hope that a model which does not make use of linguistic structure (and correspondingly spatial relations between objects) performs worse on \grefa even when it performs well on \grefh due to exploiting biases in the training data. 

Figure~\ref{tab:examples} shows several examples from the \grefe, \grefh, and \grefa splits.
We note that \grefa expressions are longer on average than \grefe and \grefh (Figure~\ref{fig:lengthdist} in appendix) and consists of rich and diverse spatial relationships (Figure~\ref{fig:datastats} in appendix). 

Concurrent to our work, \citet{gardner2020evaluating} also propose perturbed test splits for several tasks by modifying in-domain examples.
In their setup, the original authors of each task create perturbed examples, whereas we use crowdworkers.
Closest to our work is from \citet{Kaushik2020Learning} who also use crowdworkers.
While we use perturbed examples to evaluate robustness, they also use them to improve robustness (we propose complementary methods to improve robustness \secref{sec:models}).
Moreover, we are primarily concerned with the robustness of models for visual expression recognition task, while \citeauthor{gardner2020evaluating} and \citeauthor{Kaushik2020Learning} focus on different tasks (e.g., sentiment, natural language inference).

\subsection{Human Performance on \grefe, \grefh and \grefa}
We conducted an additional human study (on AMT) to compare the human performance on \grefe, \grefh and \grefa splits. First, we randomly sampled 100 referring expressions from each of the three splits. Each referring expression is then assigned to three AMT workers and are asked to select a bounding box to identify the target object. We considered a sample to be correctly annotated by humans if at least two out of three workers select the ground-truth annotation. Through this evaluation, we obtained human performance on each of the three splits Ref-Easy, Ref-Hard, and Ref-Adv as 98\%, 95\%, and 96\% respectively.

\begin{figure*}[t]
\centering
  \includegraphics[width=0.9\linewidth]{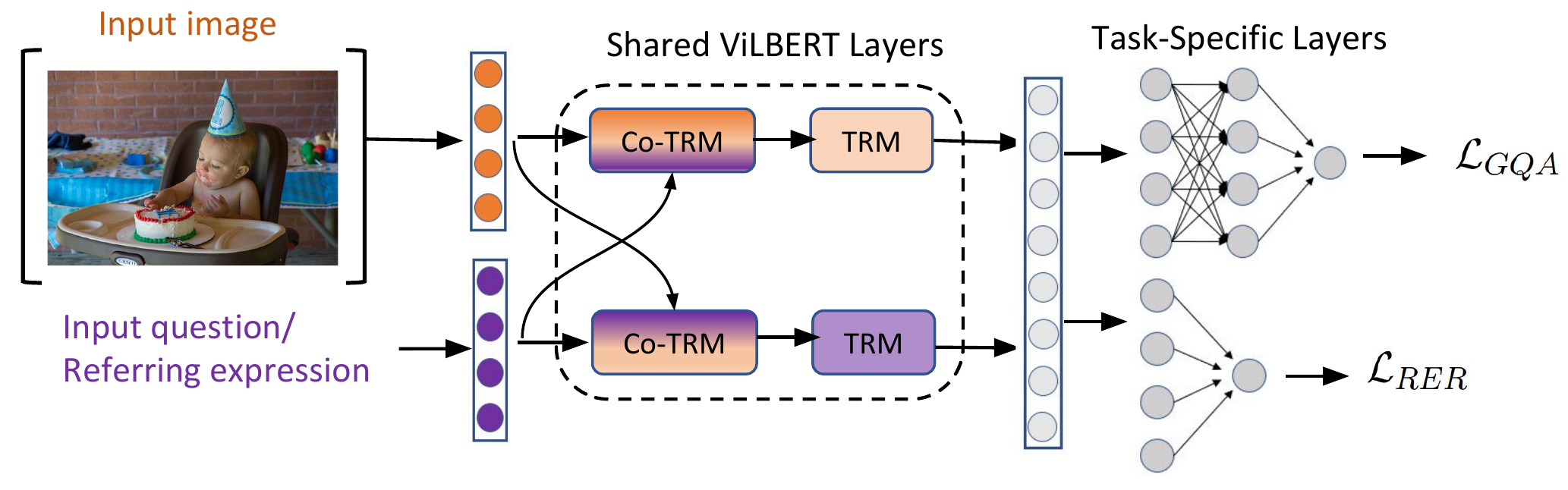}
  \vspace{-1em}
  \caption{Multi-task learning model for referring expression recognition with GQA}~\label{fig:MTL}
 \vspace{-1em}
\end{figure*}

\section{Diagnosing Referring Expression Recognition models}
\label{sec:diganising-rer}

We evaluate the following models, most of which are designed to exploit linguistic structure.

\vspace{0.25em}
\noindent\textbf{CMN}~(Compositional Modular Networks;~\citealt{hu2017modeling,andreas2016neural}) grounds expressions using neural modules by decomposing an expression into $<$subject, relation, object$>$ triples.
The subject and object are localized to the objects in the image using a localization module while the relation between them is modeled using a relationship module.
The full network learns to jointly decompose the input expression into a triple while also recognizing the target object.

\vspace{0.25em}
\noindent\textbf{GroundNet}~\cite{cirik2018using} is similar to CMN, however it makes use of rich linguistic structure (and correspondingly rich modules) as defined by an external syntactic parser.

\vspace{0.25em}
\noindent\textbf{MattNet}~\cite{yu2018mattnet} generalizes CMN to flexibly adapt to expressions that cannot be captured by the fixed template of CMN.
It introduces new modules and also uses an attention mechanism to weigh modules.

\vspace{0.25em}
\noindent\textbf{ViLBERT}~\cite{lu2019vilbert}, the state-of-the-art model for referring expression recognition, uses a \textit{pretrain-then-transfer} learning approach to jointly learn visiolinguistic representations from large-scale data and utilizes them to ground expressions. 
This is the only model that does not explicitly model compositional structure of language, but BERT-like models are shown to capture syntactic structure latently~\cite{hewitt2019structural}.

\subsection{Results and discussion}
We trained on the full training set of \gref and performed hyperparameter tuning on a development set.
We used the development and test splits of \newcite{mao2016generation}.
\tabref{tab:diagnosis} shows the model accuracies on these splits and our proposed datasets.
The models are trained to select ground truth bounding box from a set of predefined bounding boxes.
We treat a prediction as positive if the predicted bounding box has IoU $>$ 0.5 with the ground truth. %

Although the overall performance on the test set seem high, in reality, models excel only at \grefe while performing poorly on \grefh.
The difference in performance between \grefe and \grefh ranges up to 15\%. 
This indicates that current models do not exploit linguistic structure effectively.
When tested on \grefa, the performance goes down even further, increasing the gap between \grefe and \grefa (up to 26\%).
This suggests that models are relying on reasoning shortcuts found in training than actual understanding.
Among the models, GroundNet performs worse, perhaps due to its reliance on rigid structure predicted by an external parser and the mismatches between the predicted structure and spatial relations between objects.
ViLBERT achieves the highest performance and is relatively more robust than other models.
In the next section, we propose methods to further increase the robustness of ViLBERT.

\begin{table}[t]
\begin{center}
\tabcolsep 2.5pt
\begin{tabular}{lrrrrr} 
\toprule
\textit{Model} & \textit{Dev} & \textit{Test} & \textit{Easy} & \textit{Hard} & \textit{Adv} \\ 
\midrule
GroundNet %
& 66.50  & 65.80 & 67.11 & 54.47 & 42.90 \\
CMN %
& 70.00 & 69.40 & 69.55 & 68.63 & 49.50 \\
MattNet %
& 79.21 & 78.51 & 80.96 & 65.94 & 54.64 \\
ViLBERT  %
& \bf{83.39} & \bf{83.63} &  \textbf{85.93} & \textbf{72.00} & \bf{70.90} \\
\bottomrule
\end{tabular}
\caption{Accuracy of models on \gref standard splits and our splits \grefe, \grefh and \grefa.}
\label{tab:diagnosis}
\end{center}
\end{table}

\section{Increasing the robustness of ViLBERT}\label{sec:models}

We extend ViLBERT in two ways, one based on contrastive learning using negative samples, and the other based on multi-task learning on GQA \cite{hudson2019gqa}, a task that requires linguistic and spatial reasoning on images.

\paragraph{Contrastive learning using negative samples}
Instead of learning from one single example, contrastive learning aims to learn from multiple examples by comparing one to the other.
In order to increase the sensitivity to linguistic structure, we mine negative examples that are close to the current example and learn to jointly minimize the loss on the current (positive) example and maximize the loss on negative examples.
We treat the triplets $\big(i, e, b\big)$ in the training set as positive examples, where $i$, $e$, $b$ stands for image, expression and ground truth bounding box.
For each triplet $\big(i, e, b\big)$, we sample another training example $\big(i', e', b'\big)$, and use it to create two negative samples, defined by $\big(i', e, b'\big)$ and $\big(i, e', b\big)$, i.e., we pair wrong bounding boxes with wrong expressions.
For efficiency, we only consider negative pairs from the mini-batch.
We modify the batch loss function as follows:

\vspace{-1em}

\begin{equation*}
    \begin{aligned}
    \mathbf{
     \mathcal{L}\big(i, e, b\big)}& \mathbf{=} \mathbf{F_{(e,e')} \left[\ell\big(i, e, b\big) - \ell\big(i, e', b\big) - \tau\right]_{+}} \\
                       & \mathbf{+ F_{(i,i')} \left[\ell\big(i, e, b\big) - \ell\big(i', e, b'\big) - \tau\right]_{+}} \\
    \end{aligned}
\end{equation*}

Here $\ell(i, e, b)$ is the cross-entropy loss of ViLBERT, $\left[x\right]_{+}$ is the hinge loss defined by $\max\big(0,x\big)$, and $\tau$ is the margin parameter.
$F$ indicates a function over all batch samples.
We define $F$ to be either sum of hinges (Sum-H) or max of hinges (Max-H).
While Sum-H takes sum over all negative samples,
If batch size is $n$, for each $\big(i, e, b\big)$, there will be $n-1$ triplets of $\big(i', e, b'\big)$ and  $\big(i, e', b\big)$.
For $\big(i, e, b\big)$, there will be one $\big(i', e, b'\big)$ and one $\big(i, e', b\big)$.
Similar proposals are known to increase the robustness of vision and language problems like visual-semantic embeddings and image description ranking \cite{kiros2014unifying,gella-etal-2017-image,faghri2017vse++}.

\paragraph{Multi-task Learning (MTL) with GQA}
In order to increase the sensitivity to linguistic structure, we rely on tasks that require reasoning on linguistic structure and learn to perform them alongside our task.
We employ MTL with GQA~\cite{hudson2019gqa}, a compositional visual question answering dataset.
Specifically, we use the GQA-Rel split which contains questions that require reasoning on both linguistic structure and spatial relations (e.g., \textit{Is there a boy wearing a red hat standing next to yellow bus?} as opposed to \textit{Is there a boy wearing hat?}).
Figure~\ref{fig:MTL} depicts the neural architecture.
We share several layers between the tasks to enable the model to learn representations useful for both tasks.
Each shared layer constitute a co-attention transformer block (Co-TRM; ~\citealt{lu2019vilbert}) and a transformer block (TRM; \citealt{vaswani2017attention}).
While in a transformer, attention is computed using queries and keys from the same modality, in a co-attention transformer they come from different modalities (see cross arrows in Figure~\ref{fig:MTL}).
The shared representations are eventually passed as input to task-specific MLPs.
We optimize each task using alternative training \cite{luong2015multi}.

\begin{table}[t]
\begin{center}
\tabcolsep 2pt
\begin{tabular}{ p{2.6cm}ccccc}
\toprule 
\textit{Model} &  \textit{Dev} & \textit{Test} & \textit{Easy} & \textit{Hard} & \textit{Adv} \\ 
\midrule
ViLBERT (VB) & 83.39 & 83.63 & 85.93 & 72.00 & 70.90 \\

VB+\textit{Sum-H} & 81.61 & 83.00 & 85.93 & 70.60 & 72.30\\

VB+\textit{Max-H} & 82.93 & 82.70 & \bf{86.58} & 70.46 & 73.35 \\

VB+\textit{MTL (GQA)} & \bf{83.45} & \bf{84.30} & 86.23 & \textbf{73.79} & \bf{73.92} \\
\bottomrule
\end{tabular}
\caption{Accuracy of enhanced ViLBERT models.}
\label{tab:extensions}
\end{center}
\vspace{-1em}
\end{table}

\begin{figure}[t]
\centering
  \includegraphics[width=\linewidth,trim={1.8cm 1.2cm 2.8cm 1.2cm},clip]{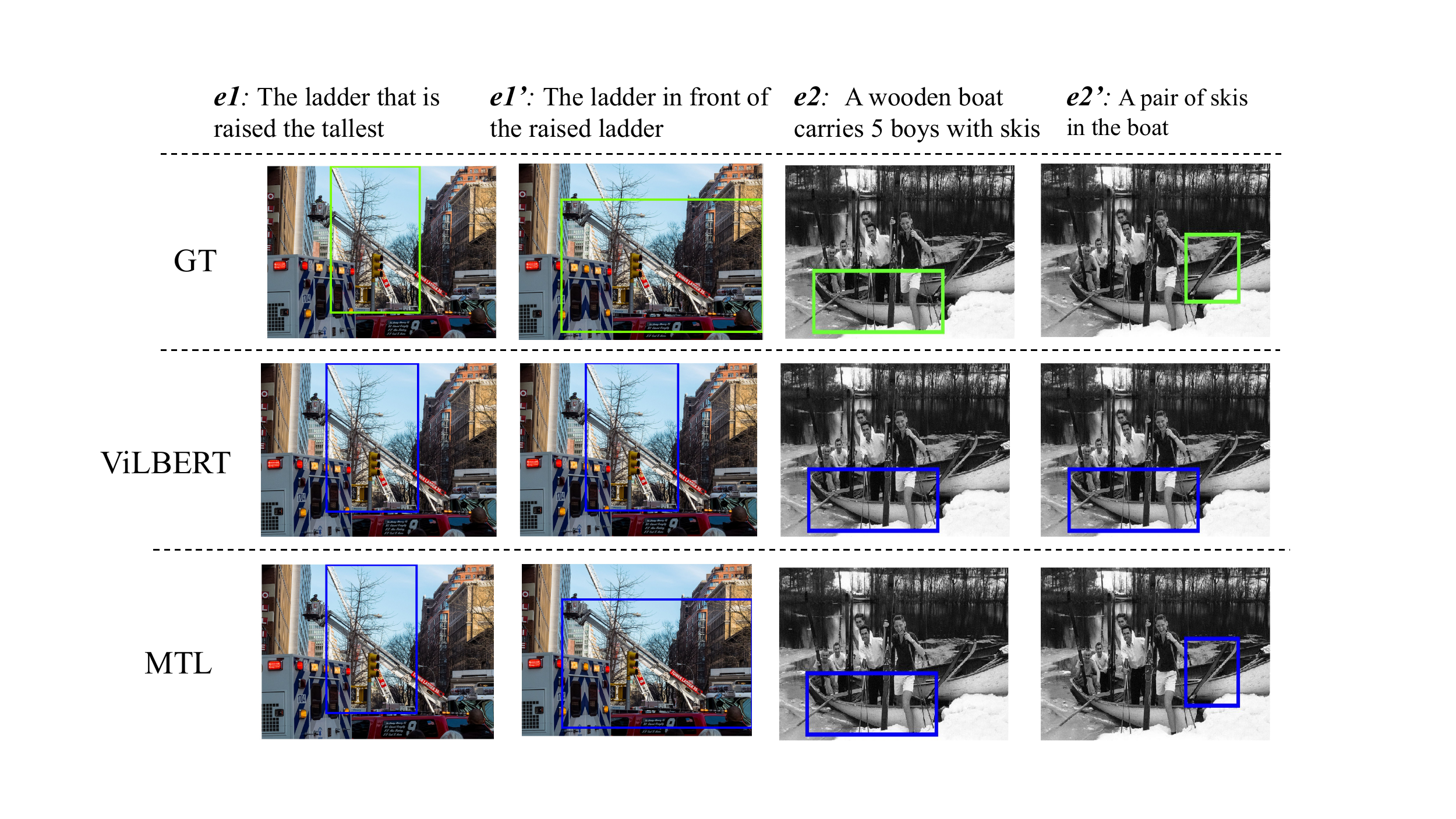}
  \caption{Predictions of ViLBERT and MTL model (GT denotes ground-truth). $e1'$ and $e2'$ are adversarial expressions of $e1$ and $e2$ respectively. }~\label{fig:qual_analysis}
  \vspace{-1em}
\end{figure}

\paragraph{Results and discussion} 
Table~\ref{tab:extensions} shows the experimental results on the referring expression recognition task.
Although contrastive learning improves the robustness of ViLBERT on \grefa (+1.4\% and +2.5\% for Sum-H and Max-H respectively), it comes at a cost of slight performance drop on the full test (likely due to sacrificing biases shared between training and test sets).
Whereas MTL improves the robustness on all sets showing that multi-task learning helps (we observe 2.3\% increase on GQA \secref{app:training}).
Moreover, the performance of MTL on \grefh and \grefa are similar, suggesting that the model generalizes to unseen data distribution.
Figure~\ref{fig:qual_analysis} shows qualitative examples comparing MTL predictions on \grefh and \grefa parallel examples.
These suggest that the MTL model is sensitive to linguistic structure.
However, there is still ample room for improvement indicated by the gap between \grefe and \grefh (12.4\%).

\section{Conclusion}\label{conclusion}
Our work shows that current datasets and models for visual referring expressions fail to make effective use of linguistic structure.
Although our proposed models are slightly more robust than existing models, there is still significant scope for improvement.
We hope that \grefh and \grefa will foster more research in this area.

\section*{Acknowledgements}
We would like to thank Volkan Cirik, Licheng Yu, Jiasen Lu for their help with GroundNet, MattNet and ViLBERT respectively, Keze Wang for his help with technical issues, and AWS AI data team for their help with Mechanical Turk.
We are grateful to the anonymous reviewers for their useful feedback.

\bibliographystyle{acl_natbib}

\begin{table}[h]
\begin{center}
\tabcolsep 2.5pt
\begin{tabular}{lrrrrr} 
\toprule
\textit{} & \greforiginal & \grefplus & \gref
  \\ 
\midrule
Outdoor %
& 0.89\% & 0.88\% & 1.65\%  \\
Food %
& 10.16\%  & 10.07\% & 8.10\%  \\
Indoor %
& 3.10\% & 3.09\% & 2.59\%  \\
Appliance  %
& 0.67\% & 0.68\% &  1.03\% \\
Kitchen  %
& 3.95\% & 3.95\% &  5.40\% \\
Accessory  %
& 2.33\% & 2.33\% &  2.85\% \\
Person  %
& 49.50\% & 49.70\% &  37.02\% \\
Animal  %
& 13.26\% & 13.27\% &  15.05\% \\
Vehicle  %
& 7.23\% & 7.22\% &  10.71\% \\
Sports  %
& 0.73\% & 0.74\% &  1.91\% \\
Electronic  %
& 1.94\% & 1.95\% &  2.56\% \\
Furniture  %
& 6.14\% & 6.12\% &  11.09\% \\
\bottomrule
\end{tabular}
\caption{Distribution of object categories in \greforiginal, \grefplus, and \gref datasets.}
\label{tab:distrib}
\end{center}
\end{table}

\newpage

\balance

\appendix

\section{Appendix}

In this supplementary material, we begin by providing more details on \gref dataset to supplement Section~\ref{sec:ling_struct} of the main paper. 
We then provide \grefa annotation details, statistics, analysis, and random examples, to supplement Section~\ref{sec:collection} of the main paper. 
Finally, we provide details of our models (initialization \& training, hyper-parameters) and show additional results to supplement Section~\ref{sec:models} of the main paper.

\subsection{\gref vs Other Referring Expressions Datasets}\label{refcocog_why}
\greforiginal, \grefplus \cite{kazemzadeh2014referitgame} and \gref (Google-RefCOCO; \citealt{mao2016generation}) are three commonly studied visual referring expression recognition datasets for real images. All the three data sets are built on top of MSCOCO dataset~\cite{lin2014microsoft} which contains more than 300,000 images, with 80 categories of objects. \greforiginal, \grefplus were collected using online interactive game. \greforiginal dataset is more biased towards person category. \grefplus does not allow the use of location words in the expressions, and therefore contains very few spatial relationships. \gref was not collected in an interactive setting and therefore contains longer expressions. %

For our adversarial analysis, we chose \gref for the following three important reasons: Firstly, expressions are longer (by 2.5 times on average) in \gref and therefore contains more spatial relationships compared to other two datasets. Secondly, \gref contains at least 2 to 4 instances of the same object type within the same image referred by an expression. This makes the dataset more robust, and indirectly puts higher importance on grounding spatial relationships in finding the target object. Finally, as shown in Table~\ref{tab:distrib}, \greforiginal and \grefplus are highly skewed towards \textit{Person} object category ($\approx$ 50\%) whereas \gref is relatively less skewed ($\approx$ 36\%), more diverse, and less biased.

\subsection{Importance of Linguistic Structure}
\newcite{CirikMB18} observed that existing models for \gref are relying heavily on the biases in the data than on linguistic structure. We perform extensive experiments to get more detailed insights into this observation. Specifically, we distort linguistic structure of referring expressions in the \gref test split and evaluate the SOTA models that are trained on original undistorted \gref training split. Similar to \cite{CirikMB18}, we distort the test split using two methods: (a) randomly shuffle words in a referring expression, and (b) delete all the words in the expression except for nouns and adjectives. Table~\ref{tab:perturb_scores1} shows accuracies for the models with (column 3 and 4) and without (column 2) distorted referring expressions. Except for the ViLBERT model\cite{lu2019vilbert}, the drop in accuracy is not significant indicating that spatial relations are ignored in grounding the referring expression.

Using the relatively robust ViLBERT model, we repeat this analysis on our splits \grefe, \grefh and \grefa. We randomly sampled 1500 expressions from each of these splits and then compare performance of ViLBERT on these three sets. As shown in Table~\ref{tab:perturb_scores2}, we find a large difference in model's accuracy on  \grefh and \grefa. This clearly indicates that grounding expressions in both of these splits require linguistic and spatial reasoning.

\subsection{\grefa Annotation}
\label{sec:annotation-steps}
\begin{figure*}[t]
\centering
  \includegraphics[width=\linewidth]{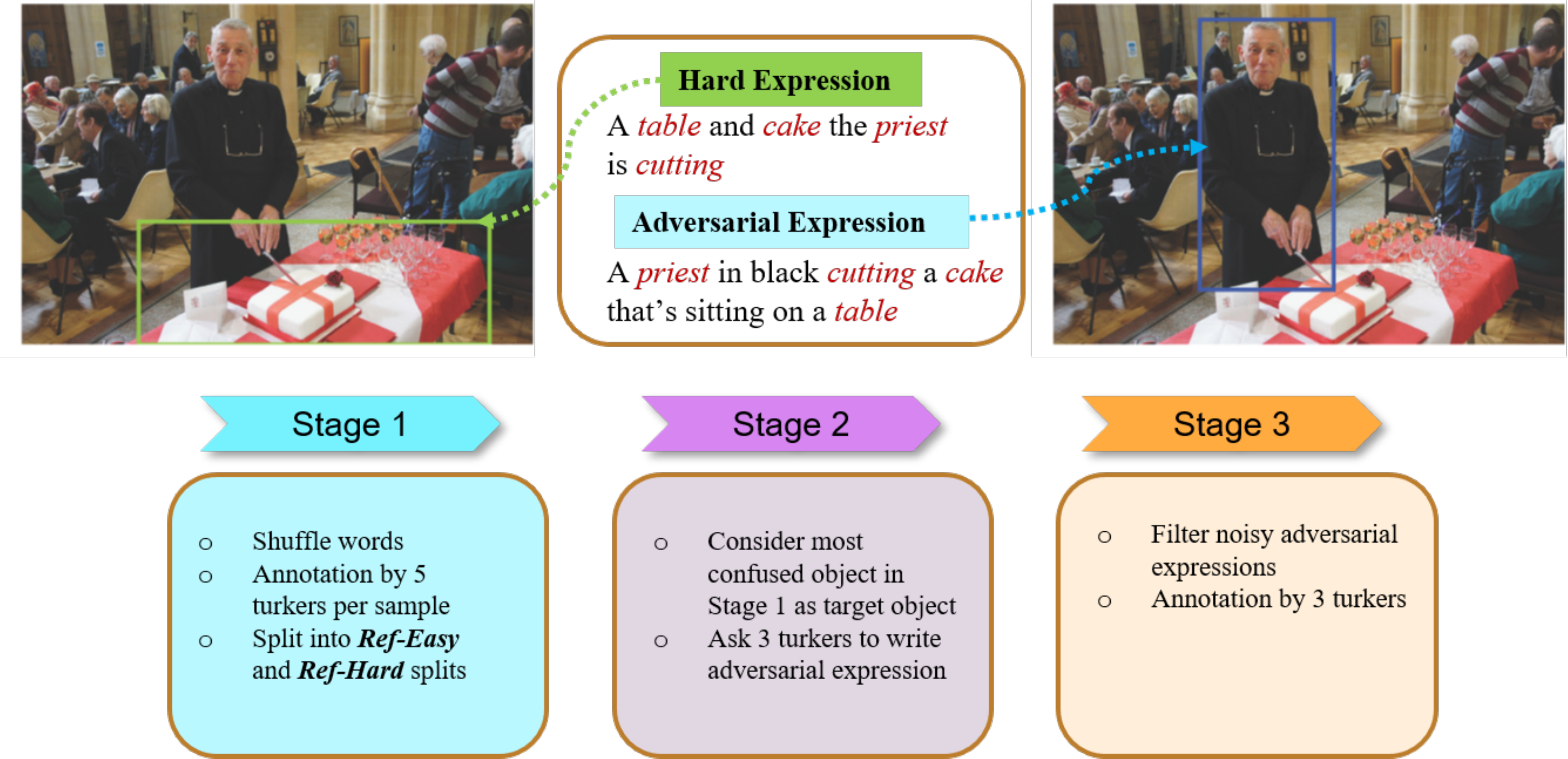}
  \caption{Overview of our three-stage \grefa construction process. Given the image, referring expression, ground-truth bounding boxes for all the samples in \gref test split, we first filter out the hard samples and then construct adversarial expressions using them. Please refer to section 2 for further detail.}~\label{fig:datacollection}
\end{figure*}

We construct \grefa by using all the 9602 referring expressions from \gref test data split. As shown in Figure~\ref{fig:datacollection}, we follow a three stage approach to collect these new samples: 

\paragraph{Stage 1:} For every referring expression in \gref test split, we perturb its linguistic structure by shuffling the word order randomly. We show each of these perturbed expression along with images and all object bounding boxes to five qualified Amazon Mechanical Turk (AMT) workers and ask them to identify the ground-truth bounding box for the shuffled referring expression. We hired workers from US and Canada with approval rates higher than 98\% and more than 1000 accepted HITs. At the beginning of the annotation, we ask the turkers to go through a familiarization phase where they become familiar with the task. %
We consider all the image and expression pairs for which at least 3 out of 5 annotators \textbf{failed to locate} the object correctly (with IoU $<$ 0.5 ) as hard samples (\textbf{\grefh}). We refer to the image-expressions for which at least 3 out of 5 annotators were \textbf{able to localize} the object correctly as easy samples (\textbf{\grefe}). On average, we found that humans failed to localize the objects correctly in 17\% of the expressions. %

\begin{table}[t]
    \centering
    \begin{tabular}{p{3.4cm} c c c}
    \toprule
         Model& Original & Shuf & N+J \\
         \midrule
        \small{CMN \cite{hu2017modeling}} & 69.4 & 66.4 & 67.4\\
        \small{GroundNet \cite{cirik2018using}}  & 65.8 & 57.6 & 62.8 \\
        \small{MattNet \cite{yu2018mattnet}}& 78.5 & 75.3 & 76.1\\
        \small{ViLBERT \cite{lu2019vilbert}} & 83.6 & 71.4 & 73.6\\
        \bottomrule
    \end{tabular}
    \caption{\gref test accuracies of SOTA models on (a) original undistorted split, (b) after randomly shuffling words (Shuf) in the referring expression, and (c) after deleting all the words except for nouns and adjectives (N+J). ViLBERT is relatively more robust than other baselines.}
    \label{tab:perturb_scores1}
\end{table}

\begin{table}[t]
    \centering
    \begin{tabular}{l c c c c}
    \toprule
        Test & Original  & Shuf  & N+J \\
         \midrule
        \grefe & 86.40 & 75.06 & 76.00 \\ 
        \grefh & 72.73 & 51.13 & 56.60 \\
        \grefa & 71.08 & 50.23 & 57.40 \\
        \bottomrule
    \end{tabular}
    \caption{\grefe, \grefh, and \grefa test accuracies of ViLBERT on (a) original undistorted split, (b) after randomly shuffling words (Shuf) in the referring expression, and (c) after deleting all the words except for nouns and adjectives (N+J).}
    \label{tab:perturb_scores2}
\end{table}

\begin{figure}[t]
\centering
  \includegraphics[width=\linewidth]{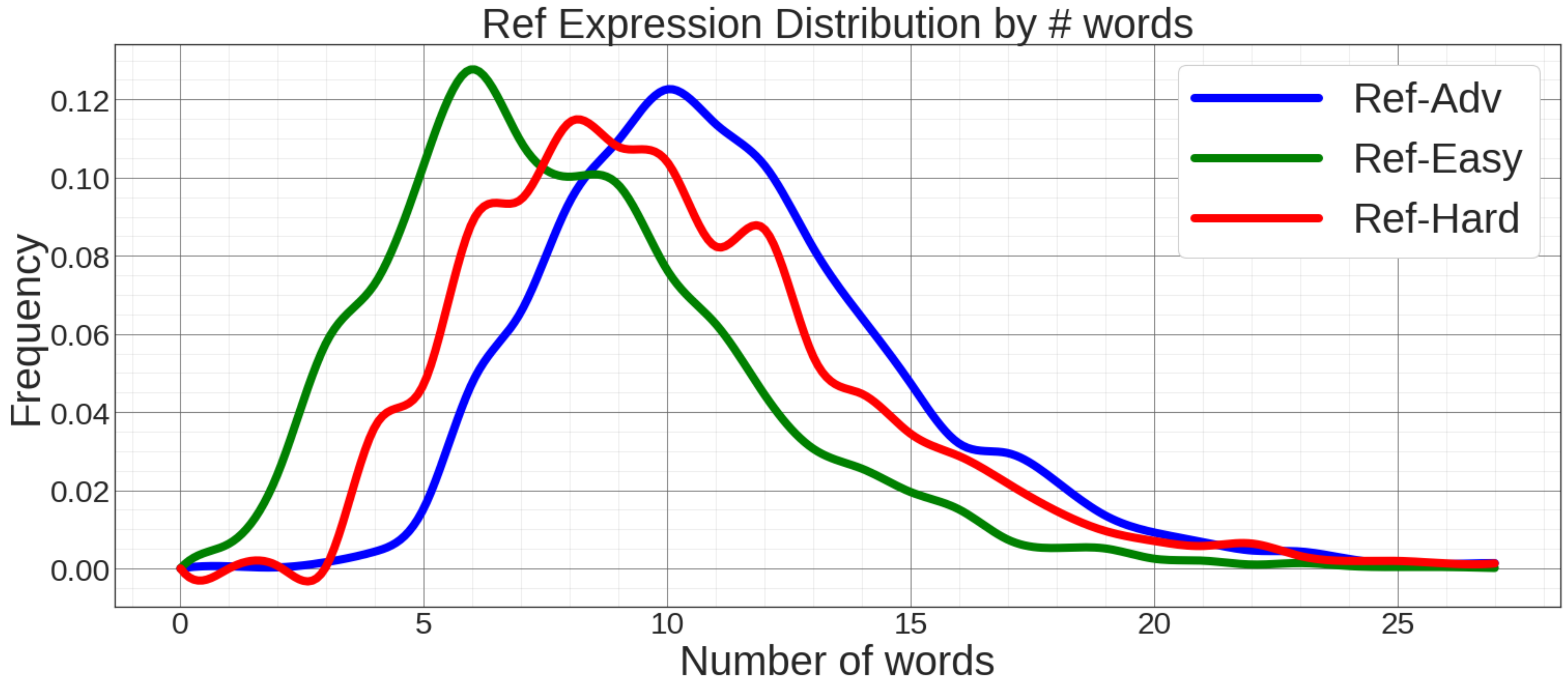}
  \caption{Referring expression length distribution for \grefe, \grefh, \grefa datasets.}
  ~\label{fig:lengthdist}
\end{figure}

\paragraph{Stage 2:} We take \grefh images and ask turkers to generate adversarial expressions such that the target object is different from the original object. More concretely, for each of the hard samples, we identify the most confused image regions among human annotators as the target objects in stage 1. For each of these target objects, we then ask three turkers to write a referring expression while retaining at least three content words (nouns and adjectives) in the original referring expression. This generates adversarial expressions for the original ground-truth \grefh referring expressions. 
 
\paragraph{Stage 3:} We filter out the noisy adversarial expressions generated in stage 2 by following a validation routine used in the generation of \gref dataset. We ask three additional AMT workers to select a bounding box to identify the target object in the adversarial expression and then remove the noisy samples for which the inter-annotator agreement among workers is low. The samples with at least 2 out of 3 annotators achieving IoU $>$ 0.5 will be added to \grefa dataset.

\begin{figure*}[t]
\centering
 \includegraphics[width=\linewidth]{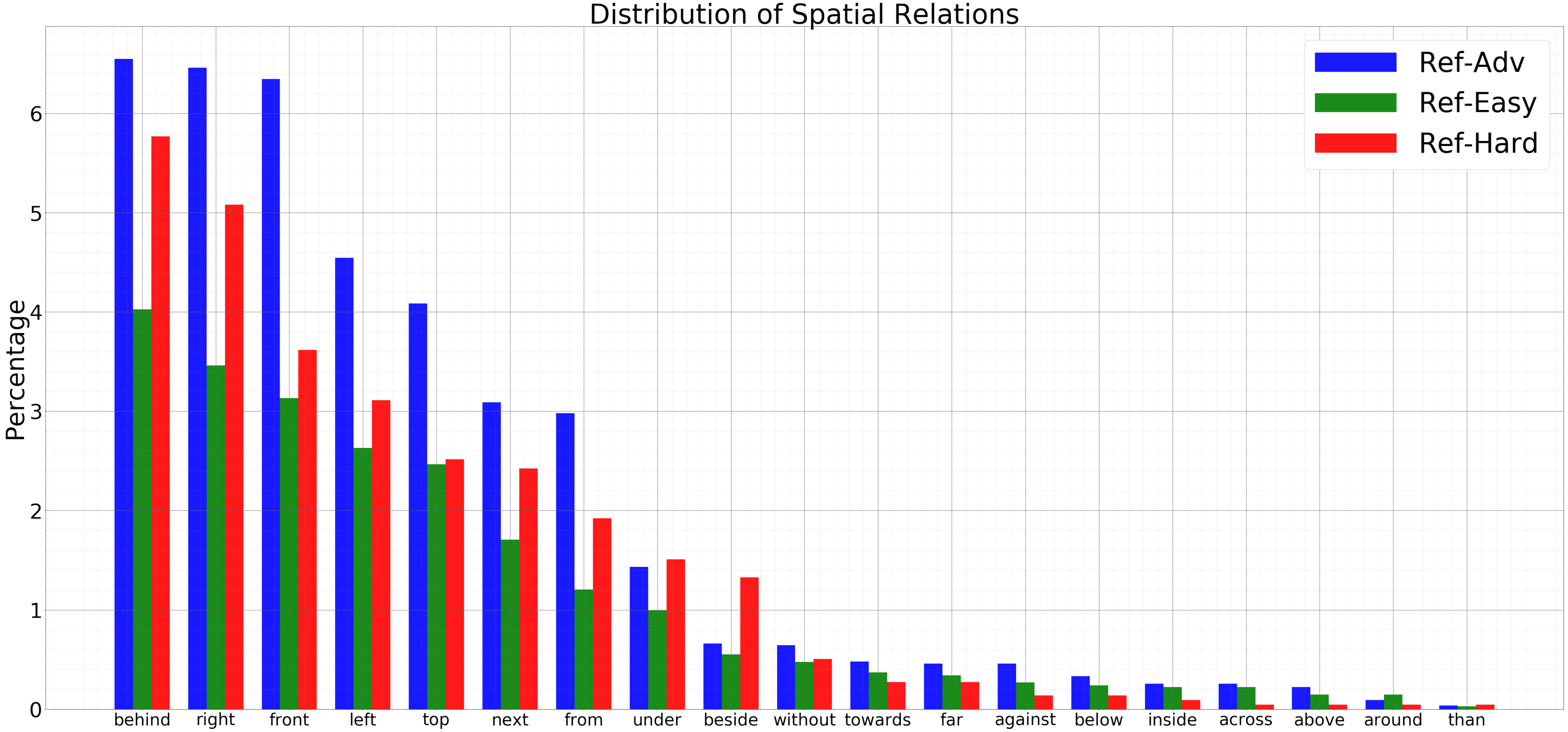}
 \caption{Relative frequency of the most frequent spatial relationships in \grefe, \grefh, and \grefa}~\label{fig:datastats}
\end{figure*}

\begin{table}[t]
\begin{center}
\tabcolsep 2pt
\begin{tabular}{p{5cm}r  } 
\hline
Referring Expressions & 3704\\
Unique Images & 976   \\
Vocabulary & 2319\\
Avg. Length of Expression &  11.4\\
\hline
\end{tabular}
\caption{\grefa Statistics}
\label{tab:app-stats}
\end{center}
\end{table}

\begin{table}[t]
\begin{center}
\tabcolsep 2.5pt
\begin{tabular}{lrrrrr}
\toprule
\textit{} & $\underset{8034 \textrm{ samples}}{\grefe}$ & $\underset{1568 \textrm{ samples}}{\grefh}$ & $\underset{3704 \textrm{ samples}}{\grefa}$ \\ 
\midrule
Outdoor %
& 1.21\% & 1.90\% & 1.97\%  \\
Food %
& 7.94\%  & 9.80\% & 9.63\%  \\
Indoor %
& 2.81\% & 2.83\% & 2.76\%  \\
Appliance  %
& 0.80\% & 1.07\% &  1.11\% \\
Kitchen  %
& 4.52\% & 5.73\% &  5.77\% \\
Accessory  %
& 3.20\% & 5.44\% &  5.29\% \\
Person  %
& 37.26\% & 20.88\% & 21.01\% \\
Animal  %
& 15.95\% & 13.92\% & 13.90\% \\
Vehicle  %
& 10.91\% & 10.40\% & 10.26\% \\
Sports  %
& 1.45\% & 5.04\% &  5.13\% \\
Electronic  %
& 2.62\% & 3.20\% &  3.31\% \\
Furniture  %
& 11.28\% & 19.73\% &  19.83\% \\
\bottomrule
\end{tabular}
\caption{Distribution of object categories in \grefe, \grefh, and \grefa splits.}
\label{tab:ref_splits_distrib}
\end{center}
\end{table}

\subsection{Dataset Analysis, Comparison, and Visualization}
In Table~\ref{tab:app-stats} we summarize the size and complexity of our \grefa split. Figure~\ref{fig:lengthdist} shows expression length distribution of \grefe, \grefh, and \grefa. It should be noted that \grefa expressions are longer on average than \grefe and \grefh. Distribution of object categories in \grefe, \grefh and \grefa is shown in Table ~\ref{tab:ref_splits_distrib}. In comparison to \grefe and \grefh, \grefa is more balanced and less biased towards \texttt{Person} category. Figure~\ref{fig:datastats} shows the relative frequency of the most frequent spatial relationships in all the three splits. As we can see, \grefa comprises of rich and diverse spatial relationships. In Table~\ref{tab:examples}, we show random selection of the \grefe, \grefh, and \grefa splits.

\noindent

\subsection{Model and other Experiment Details}
\subsubsection{Datasets}
\textbf{GQA}~\cite{hudson2019gqa} contains 22M questions generated from Visual Genome~\cite{krishna2017visual} scene graphs. However, in our our multi-task training (MTL), we leverage only 1.42M questions that require reasoning on both linguistic structure and spatial relations. %
We filter these relational questions by applying the following constraint on question types: \textit{type.Semantic=`\textbf{rel}'}. We also apply this constraint for filtering the development set.
We denote this subset as \textit{GQA-Rel}. %
We considered GQA-Rel instead of GQA for two reasons: 1) GQA-Rel is a more related task to RefCOCOg; and 2) MTL training with the full GQA set is computationally expensive.
For each question in the dataset, there exists a long answer (free-form text) and a short answer (containing one or two words). We only consider the short answers for the questions and treat the unique set of answers as output categories. While the full GQA dataset has 3129 output categories, GQA-Rel contains only 1842 categories. 

We follow \citet{yu2018mattnet} in creating the train (80512 expressions), val (4896 expressions), and test (9602 expressions) splits of \textbf{\gref}. For all our experiments in this paper, we directly use the ground-truth bounding box proposals.

\subsubsection{Training}
\label{app:training}

\paragraph{ViLBERT Pre-training} 
 We used pre-trained ViLBERT model that is trained on 3.3 million image-caption pairs from Conceptual Captions dataset~\cite{sharma2018conceptual}.\footnote{ViLBERT 8-Layer model at the link \url{https://github.com/jiasenlu/vilbert_beta}}

\paragraph{Single-Task Fine-tuning on \gref} In order to fine-tune the baseline ViLBERT~\cite{lu2019vilbert} model on \gref dataset, we pass the ViLBERT visual representation
for each bounding box into a linear layer to predict a matching score (similar to RefCOCO+ training in \citealt{lu2019vilbert}). We calculate accuracy using IoU metric (prediction is correct if IoU(predicted\_region, ground-truth region) $>$ 0.5). We use a binary cross-entropy loss and train the model for a maximum of 25 epochs. We use early-stopping based on the validation performance. We use an initial learning rate of 4e-5 and use a linear decay learning rate schedule with warm up. We train on 8 Tesla V100 GPUs with a total batch size of 512.

\paragraph{Negative Mining}
We used a batch size of 512 and randomly sample negatives from the mini-batch for computational efficiency. We sampled 64 negatives from each batch for both Sum of Hinges and Max of Hinges losses. We fine-tune the margin parameters based on development split. We train the model for a maximum of 25 epochs. We use early-stopping based on the validation performance. We use an initial learning rate of 4e-5 and use a linear decay learning rate schedule with warm up. We train on 8 Tesla V100 GPUs with a total batch size of 512.

\begin{table}[tp]
    \centering
    \tabcolsep 2pt
    \begin{tabular}{lcc}
    \toprule
     Split & Before MTL & After MTL \\
    \midrule
  GQA-Rel Dev & 53.7\% & 56.0\% \\
 GQA Dev & 40.24\% &  42.1\% \\
 GQA Test & 36.64\% &  39.2\% \\
\bottomrule
    \end{tabular}
    \caption{Performance on GQA-Rel Dev, GQA-Dev and GQA-Test splits \textit{before} and \textit{after} MTL training with \gref (Note: MTL training for all the three rows is performed using GQA-Rel and \gref). 
    }
    \label{tab:perf_gqa}
\end{table}

\paragraph{Multi-Task Learning (MTL) with GQA-Rel}
The multi-task learning architecture is shown in Figure~\ref{fig:MTL} in the main paper. The shared layers constitute transformer blocks (TRM) and co-attentional transformer layers (Co-TRM) in ViLBERT \cite{lu2019vilbert}. The task-specific layer for GQA task is a two-layer MLP and we treat it as a multi-class classification task and the task-specific layer for RER is a linear layer that predicts a matching score for each of the image regions given an input referring expression. The weights for the task-specific layers are randomly initialized, whereas the shared layers are initialized with weights pre-trained on 3.3 million image-caption pairs from Conceptual Captions dataset~\cite{sharma2018conceptual}. We use a binary cross-entropy loss for both tasks. Similar to \citet{luong2015multi}, during training, we optimize each task alternatively in mini-batches based on a mixing ratio. We use early-stopping based on the validation performance. We use an initial learning rate of 4e-5 for \gref and 2e-5 for GQA, and use a linear decay learning rate schedule with warm up. We train on 4 RTX 2080 GPUs with a total batch size of 256.

\paragraph{GQA MTL Results} Table 3 in the main paper showed that MTL training with GQA-Rel significantly improved the performance of model on \grefh and \grefa splits. In addition, we also observed a significant improvement in GQA-Rel development, GQA development and test splits as shown in the Table~\ref{tab:perf_gqa}.

\subsubsection{Additional Experiments}

In this subsection, we present results of additional experiments using transfer learning (TL) and multi-task learning (MTL) with ViLBERT on VQA, GQA, and GQA-Rel tasks. As shown in Table~\ref{tab:transfer_learning}, TL with VQA showed slight improvement. However, TL with GQA, TL with GQA-Rel, and MTL with VQA did not show any improvements~\footnote{We could not perform MTL with GQA as it requires large number of computational resources.}.

\begin{table}[h]
    \centering
    \tabcolsep 2pt
    \begin{tabular}{p{3.4cm} c c c}
    \toprule
         ViLBERT & \grefd & \greft & \grefa \\
         \midrule
        Without TL and MTL & 83.39 & 83.63 & 70.90 \\
        TL with VQA & 82.26 & 84.14 & 72.96 \\ 
        TL with GQA & 80.60 & 82.08 & 70.41 \\
        TL with GQA-Rel & 81.05 & 83.12 &  70.78\\
        MTL with VQA & 81.20 & 82.10 & 70.82 \\
        MTL with GQA-Rel & \textbf{83.45} & \textbf{84.30} & \textbf{73.92} \\
        \bottomrule
    \end{tabular}
    \caption{Comparing ViLBERT's Multi-task Learning (MTL) with Transfer Learning (TL) experiments. \grefd and \greft correspond to: \grefDev and \greftest splits respectively.}
    \label{tab:transfer_learning}
\end{table}

\end{document}